\documentclass[pdflatex,sn-basic]{sn-jnl}

\usepackage{graphicx}%
\usepackage{multirow}%
\usepackage{amsmath,amssymb,amsfonts}%
\usepackage{amsthm}%
\usepackage{mathrsfs}%
\usepackage[T1]{fontenc}
\usepackage[title]{appendix}%
\usepackage{xcolor}%
\usepackage{textcomp}%
\usepackage{manyfoot}%
\usepackage{booktabs}%
\usepackage{chemfig}
\usepackage{latexsym}
\usepackage{algorithm}%
\usepackage{enumitem}
\usepackage{algpseudocode}%
\usepackage{listings}%
\usepackage{subcaption}
\usepackage{flushend}
\usepackage{tikz}
\usetikzlibrary{arrows, decorations.pathmorphing,backgrounds, positioning,fit,petri,tikzmark}
\usepackage{float}
\usepackage{lscape}

\usepackage{pgfplots}
\usepgfplotslibrary{polar}
\pgfplotsset{compat=1.18}

\tikzstyle{arrow} = [thick,->,>=stealth]
\tikzstyle{process} = [rectangle,rounded corners, thick, minimum width=.6cm, minimum height=.5cm, text centered, draw=black]

\newtheorem{definition}{Definition}
\newtheorem{rules}{Rule}
\hypersetup{hidelinks}

\theoremstyle{thmstyleone}%
\theoremstyle{thmstyletwo}%

\theoremstyle{thmstylethree}%

\newcommand{\atom}[2]{$\langle\textit{{#1}},\textit{#2}\rangle$}

\begin{document}

\title[Article Title] {Execution Flexibility in Automated Planning: A Comparative Evaluation of Deordering and Reordering Strategies }


\author[1]{\fnm{Md. Monjurul} \sur{Islam}}\email{monjurul.islam.cs@gmail.com}

\author[1]{\fnm{Sabah Binte} \sur{Noor}}\email{sabah@duet.ac.bd}

\author[1]{\fnm{Fazlul Hasan} \sur{Siddiqui}}\email{siddiqui@duet.ac.bd}

\author*[2]{\fnm{Gahangir} \sur{Hossain}}\email{Gahangir.Hossain@unt.edu}

\affil[1]{\orgdiv{Department of Computer Science and Engineering}, \orgname{Dhaka University of Engineering \& Technology}, \orgaddress{\city{Gazipur}, \postcode{Gazipur-1707}, \country{Bangladesh}}}

\affil[2]{\orgdiv{Department of Data Science}, \orgname{, University of North Texas, Denton, TX}, \orgaddress{\country{USA}}}

\abstract{
This study covers foundational concepts for enhancing plan-execution flexibility, including partial-order planning, the producer-consumer-threat formalism, and a range of deordering and reordering strategies. Creating a partial-order plan from a sequential one by removing unnecessary ordering constraints is a practical way to improve execution flexibility, and several methods have been proposed for this task. This study analyzes their capabilities across ordering, action handling, parameter handling, plan structure, concurrency, and complexity, and evaluates them against each other on a shared benchmark. The central finding is that block deordering-based approaches, which restructure causal dependencies through block-level grouping and subplan substitution, substantially outperform MaxSAT-based approaches despite the latter's theoretical guarantees of minimum reordering. The reason is structural: minimum reordering optimizes within the causal structure already present in the plan, whereas block deordering-based methods change that structure, exposing orderings that would otherwise appear necessary. A further distinction is practical: block deordering-based methods are anytime algorithms that always return a valid result, while MaxSAT-based methods fail entirely on a substantial portion of plans and offer no partial solution when they do. Block substitution further extends the parallel execution by formalizing non-concurrency constraints, though its impact is limited to domains with resource-based interactions. On efficiency, block deordering-based approaches achieve the highest flex gain per unit of computation time, while MaxSAT-based encodings incur large computational overhead.
}
\keywords{Automated Planning, Flexibility, Concurrency, Partial Order Planning}

\maketitle

\section{Introduction}
\label{introduction}
The capability of autonomous agents to adapt to unforeseen circumstances represents a fundamental requirement for effective operation in dynamic environments. Within the domain of automated planning, multiple methodologies have been developed to enhance agent flexibility, ranging from providing agents with alternative plan selections \citep{1decaf} to extending the applicability of existing plans through generalization techniques \citep{Anderson1988}. 
Plan generalization approaches generate partial-order plans (POPs) by reprieving specific action orderings until such orderings become inevitable.
This strategy of minimizing action ordering decisions has been explored through various concepts, including plan deordering \citep{Kambhampati1, Veloso2002,siddiqui_patrik_2012} and reordering \citep{maxsat} methodologies.
While deordering the plan removes inessential action orderings, reordering the plan permits arbitrary changes to the action sequence.

One way to produce a POP is through partial-order causal link (POCL) planning \citep{weld_1994}, which constructs a plan incrementally while maintaining causal support for every action precondition. A practical alternative is to take a sequential plan from a fast heuristic planner and remove the ordering constraints that are not necessary.

Several methods have been developed for this post-processing task. Explanation-based order generalization (EOG) \citep{kk,Veloso2002} annotates each ordering with its causal justification and removes those without one. Block deordering (BD) \citep{siddiqui_patrik_2012} groups coherent operators into blocks and eliminates additional orderings by treating each block as a unit. Minimum reordering (MR) \citep{maxsat} encodes the deordering problem as partial weighted MaxSAT to find provably minimum reorderings and remove redundant actions; minimum reinstantiated reordering (MRR) \citep{maxsat_reinst} extends this with parameter rebinding within same-name operators. More recently, FIBS \citep{noor2024improving} replaces subplans with alternatives using block-substitution, and CIBS \citep{noor2025improving} extends FIBS to model and minimize non-concurrency constraints for parallel execution.

We analyze capabilities of different algorithms across multiple dimensions (ordering, action handling, parameter handling, plan structure, concurrency, and complexity), and evaluate on 3,345 plans from 46 IPC domains.
The analysis produces a result that is not obvious from the descriptions above. MR and MRR are designed to find minimum reorderings. Yet BD and FIBS produce substantially higher flex values on the same benchmarks, with large and statistically significant effect sizes against all other methods. The reason is that minimum reordering optimizes within the causal structure the original plan already has. Block deordering works differently: it reorganizes which operators share causal dependencies by grouping them into blocks, exposing ordering constraints that would otherwise appear necessary. Block-substitution goes further, replacing subplans with alternatives that carry different causal entailments entirely. The two families of methods are not in competition so much as they are solving different problems; restructuring causal structure turns out to be the more productive one for increasing flexibility.

\section{Preliminaries}
This section outlines the foundational concepts required to comprehend the methods. It begins with defining planning tasks and plans within the finite-domain representation, followed by notions of partial-order planning, causal dependencies, and threats. Finally, it outlines key concepts of plan deordering and reordering.
\subsection{Planning Task}
Classical planning problems are commonly represented using finite domain representation (FDR) \citep{fd}, which describes planning tasks through state variables and their associated value domains. The following definitions establish the mathematical foundation for subsequent analysis.

\begin{definition}
\label{def:sas}
In FDR, a planning task is denoted as $\Pi = \langle \mathcal{X}, \mathcal{O}, s_i, s_g \rangle$, where:
\begin{itemize}
    \item $\mathcal{X}$ represents a finite non-empty set of \textbf{state variables}. 
    A domain $\mathcal{D}_x$ with a finite number of elements is associated with each $x \in \mathcal{X}$. 
    The pair $\langle x,d\rangle$ with $x\in \mathcal{X}$ and $d\in \mathcal{D}_x$ is a \textbf{fact}. 
    The notion of \textbf{state} corresponds to a function \textbf{s} defined over $\mathcal{X}$, such that $s(x) \in \mathcal{D}_x$ for all $x \in \mathcal{X}$. 
    The set of variables involved in a function is denoted as $\mathit{vars}(s)$.
    A function ${\hat{s}}$ similar to a state \textbf{s}, but defined over a proper subset of $\mathcal{X}$, represents a \textbf{partial state} where $\mathit{vars}(\hat{s}) \subset \mathcal{X}$.
    \item $\mathcal{O}$ represents the finite non-empty set of \textbf{operators}. Every operator $o \in \mathcal{O}$ is associated with two partial states, \textbf{precondition} ($\mathit{pre_o}$) and \textbf{effect} ($\mathit{eff_o}$), along with an associated \textbf{cost} that is nonnegative ($\mathit{cost_o} \in \mathbb{R}^+_0$).
    An operator $o$ can be applied in a state $s$ iff $\mathit{pre_o} \subseteq s$.
    Applying the operator $o$ within state $s$ transforms it into state $s' = apply(s, o)$ as specified in \eqref{eq:transform}.  
    \begin{equation}
        s'(x) = apply(s, o) = \begin{cases} 
       \mathit{eff_o}(x) & \text{if } x \in \mathit{vars(eff_o)} \\
       s(x) & \text{otherwise}
        \end{cases}
        \label{eq:transform}
    \end{equation}
    \item $s_i$ corresponds to the \textbf{initial state}, 
    \item $s_g$ refers to a partial state that specifies the \textbf{goal conditions}.
\end{itemize}
\end{definition}

\begin{definition}
    Let $\pi = \langle o_1, o_2, \dots, o_k, \dots, o_n \rangle$ be a \textbf{plan}, a sequence of operators, for a planning task 
    $\Pi = \langle \mathcal{X}, \mathcal{O}, s_i, s_g \rangle$. 
    The plan $\pi$ is \textbf{valid} iff $pre_{o_1} \subseteq s_i$, $pre_{o_{k+1}} \subseteq s_k$ for every $k \in \{1,2,\dots,n-1\}$ where $s_k = \mathit{apply}(s_{k-1}, o_k)$, and $s_g \subseteq s_n$.
    
\end{definition}

\begin{definition}
\label{def:op}
The \textbf{prod\textsubscript{o}}, \textbf{cons\textsubscript{o}}, and \textbf{del\textsubscript{o}} represent the facts that an operator $o$ produces, consumes, and deletes, respectively. The fact$-$
    \begin{itemize}
        \item $\langle x, d \rangle\in prod_o$ iff $\langle x, d \rangle \in \mathit{eff_o}$.    
        \item $\langle x, d \rangle\in cons_o$ iff $\langle x, d \rangle \in \mathit{pre_o}$.
        \item $\langle x, d \rangle\in del_o$ iff-
        \begin{enumerate}[label=\roman*.]
            \item  either  $x \notin vars(\mathit{cons_o})$ or $\mathit{cons_o}(x) =  d$,  and 
            \item  $\mathit{eff_o}(x) = d'$ such that $\ d' \in (\mathcal{D}_x\setminus \{d\})$.  
        \end{enumerate}
    \end{itemize}
\end{definition}

\subsection{Partial-Order Planning}
Operators can be executed in any possible sequence using the partial-order plan (POP) framework while imposing partial orderings over them. It is assumed that each operator is individually identifiable, even if an operator may appear multiple times within a POP.
\begin{definition}\label{def:pop}
A \textbf{POP} is represented as $\pi_{pop} =\langle \mathcal{O}, \prec \rangle$, regarding a planning task $\Pi=\langle \mathcal{X}, \mathcal{O}, s_i, s_g\rangle$, where:
\begin{itemize}
    \item $\mathcal{O}$ denotes the set of operators, and
    \item $\prec$ represents ordering constraints over $\mathcal{O}$. The notation $\mathbf{o_a \prec o_b}$ represents a \textbf{ordering constraint} between two operators $o_a, o_b \in \mathcal{O}$, which specifies that the operator $o_a$ needs to be executed at any time before the execution of operator $o_b$. $\prec$ exhibits transitive property, which means that if $o_a \prec o_b$ and $o_b \prec o_c$, then $o_a \prec o_c$. Basic orderings represent constraints that are not transitively inferred by other constraints in the set.
\end{itemize}
\end{definition} 

Plan flexibility quantification uses the metric $\mathit{flex}$ \citep{siddiqui_patrik_2012}, which is the proportion of the pairs of operators that do not have fundamental or transitive ordering to all operator pairs. Higher $\mathit{flex}$ values indicate greater execution flexibility, though the relationship between flexibility and practical adaptability depends on domain-specific factors.
\begin{definition}
    Let $\pi_{pop}=\langle \mathcal{O}, \prec\rangle$ be a POP. The \textbf{flex} value of the plan $\pi_{pop}$ is $\mathit{flex}(\pi_{pop})$ and is calculated as specified in \eqref{eq:flex}.
    \begin{equation}
        \mathit{flex}(\pi_{pop}) = 1- \frac{|\prec|}{\Sigma_{i=1}^{|\mathcal{O}|-1}i}
        \label{eq:flex}
    \end{equation}
    where $\Sigma_{i=1}^{|\mathcal{O}|-1}i$ is the maximum number of pairings that might be constructed with a collection of $|\mathcal{O}|$ items.
\end{definition}

The framework of producer-consumer-threat (PCT) \citep{backstrom1998} establishes ordering structures by identifying operator relationships regarding fact production, consumption, and deletion. This formalism introduces causal links that map operator preconditions to effect providers.


\begin{definition}
A \textbf{causal link} $o_p \xrightarrow{\langle x, d \rangle} o_c$ between $o_p$ and $o_c$ specifies that  $o_p \prec o_c$ and the operator $o_p$  produces $\langle x, d \rangle$ for the operator $o_c$, where $\langle x, d \rangle\in (prod_{o_p} \cap cons_{o_c})$.
A \textbf{threat} is defined as a conflict between a causal link $o_p \xrightarrow{\langle x, d \rangle} o_c$ and the effect of an operator $o_d$, where $o_d$ deletes $\langle x, d \rangle$ and can be ordered in between $o_p$ and $o_c$.
\end{definition}

A \textbf{promotion}, which adds the ordering $o_d \prec o_p$, or a \textbf{demotion}, which adds the ordering $o_c \prec o_d$, can resolve a threat between a causal link $o_p \xrightarrow{\langle x, d \rangle} o_c$ and an operator $o_d$. A POP is valid if causal relationships support each operator precondition without any threat \citep{weld_1994}.  Three labels, $PC, CT$, and $TP$, are introduced by \cite{siddiqui_patrik_2012} to outline a POP's ordering constraints. 

\begin{definition}
Let $\pi_{pop} =\langle \mathcal{O}, \prec \rangle$ be a POP and $Re(o_a \prec o_b)$ represents the ordering reasons for an ordering constraint $o_a \prec o_b$.
An ordering restriction $o_a \prec o_b$ might form for three reasons:
    \begin{itemize}
    \item $PC(\langle x, d \rangle) \in Re(o_a \prec o_b)$ is the \textbf{producer-consumer} of a fact \atom{x}{d}, where $o_a$ produces $\langle x, d \rangle$ and $o_b$ consumes $\langle x, d \rangle$.  The fact $\langle x, d \rangle$ may be produced and consumed by multiple operators. One operator $o_a$ is assigned via a causal link to obtain $\langle x, d \rangle$ for $o_b$.

    \item $CT(\langle x, d \rangle) \in Re(o_a \prec o_b)$ is the \textbf{consumer-threat} of a fact \atom{x}{d}, where operator $o_a$ consumes $\langle x, d \rangle$ and $o_b$ deletes $\langle x, d \rangle$.

    \item $TP(\langle x, d \rangle) \in Re(o_a \prec o_b)$ is the \textbf{threat-producer} of a fact \atom{x}{d}, where operator $o_a$ deletes the fact $\langle x, d \rangle$ and there exist one or more causal link $o_b \xrightarrow{\langle x, d \rangle} o_c$ for some $o_c\in \mathcal{O}$.
\end{itemize}
\label{def:pc-cd-dp}
\end{definition}
Causal links of a POP are denoted by the label $PC$, and the demotion and promotion ordering constraints of the plan are denoted by $CT$ and $TP$, respectively. These labels assist in identifying and monitoring the rationale behind the orderings in a POP.

\subsection{Plan Deordering and Reordering}
Plan \emph{deordering} and \emph{reordering} are two key concepts for attaining plan execution flexibility.
Formal definitions of these notions are given in \cite{backstrom1998}. A POP is assumed to be transitively closed in the Definition \ref{def:de_re}.
 \begin{definition}
 \label{def:de_re}
Let $\pi_p=\langle \mathcal{O}, \prec_p \rangle$ and $\pi_q=\langle \mathcal{O}, \prec_q \rangle$ be two different POPs with respect to $\Pi$ (a planning task). Therefore:
    \begin{itemize}
         \item $\pi_q$ is a \textbf{deordering} of $\pi_p$  with respect to $\Pi$ iff both $\pi_p$ and $\pi_q$ are valid POP and $\prec_q \subseteq \prec_p$.
         \item $\pi_q$ is a \textbf{proper deordering} of $\pi_p$  with respect to $\Pi$ iff $\pi_q$ is a deordering of $\pi_p$ and $\prec_q \subset \prec_p$.
          \item $\pi_q$ is a \textbf{reordering} of $\pi_p$  with respect to $\Pi$ iff both $\pi_p$ and $\pi_q$ are valid POP.
          \item $\pi_q$ is a \textbf{proper reordering} of $\pi_p$  with respect to $\Pi$ iff $\pi_q$ is a reordering of $\pi_p$ with $\prec_q \ne \prec_p$.
         \item $\pi_q$ is a \textbf{minimum deordering} of $\pi_p$  with respect to $\Pi$ iff$-$
         \begin{enumerate}
             \item $\pi_q$ is a deordering of $\pi_p$, and
             \item there exists no partial-order plan $\pi_r=\langle \mathcal{O}, \prec_r\rangle$ such that $\pi_r$ is a deordering of $\pi_p$ with $|\prec_r|<|\prec_q|$.
         \end{enumerate}
         \item $\pi_q$ is a \textbf{minimum reordering} of $\pi_p$  with respect to $\Pi$ iff$-$
         \begin{enumerate}
             \item $\pi_q$ is a reordering of $\pi_p$, and
             \item there exists no partial-order plan $\pi_r=\langle \mathcal{O}, \prec_r\rangle$ such that $\pi_r$ is a reordering of $\pi_p$ and $|\prec_r| < |\prec_q|$.
         \end{enumerate}
    \end{itemize}
 \end{definition}

\section{Partial-Order Causal Link-based Approach}
The partial-order causal link (POCL) \citep{weld_1994} planning framework serves as the foundation for the traditional method of creating a partial-order plan (POP).
An initial POP is iteratively refined through modifications involving operators, causal links, and ordering constraints in this paradigm. Such a preliminary plan typically begins with an initial operator and a goal operator. 
Refinement steps may introduce a new operator, insert a causal link connecting two operators, or establish an ordering constraint between them. 
A POP is considered completed once each operator's precondition is protected by a threat-free causal link. 
The two most popular POCL-based partial-order planners are VHPOP \citep{vhpop} and UCPOP \citep{ucpop}. Furthermore, the POCL strategy has been adapted to find temporal plans through state-based forward search techniques \citep{Coles_Coles_Fox_Long_2021}, and it also forms the foundation for numerous hierarchical planning approaches \citep{bercher_ijcai2017p68,Bercher_2016,bitmonnot2020fape,bitmonnot:hal-01319768}.

In addition to POCL, other techniques have been developed to generate POPs. One such method is \emph{Petri net unfolding} \citep{petri_net}, which uses iterative unfolding of a specifically constructed Petri net to simulate the execution of a forward planning system.
Another prominent approach is \emph{Graphplan} \citep{BLUM1997281}, which leverages a compact structure called a \emph{planning graph} to derive optimal POPs. Over time, this planning graph has been used as a preprocessing mechanism for diverse planning systems, including STAN \citep{STAN}, IPP \citep{IPP}, and Blackbox \citep{blackbox}.
\section{Partial Weighted MaxSAT-based Approach}
Transforming sequential plans into partial-order plans through deordering or reordering is a prominent approach for generating POPs. Early strategies for deordering involved generalizing and storing sequential plans in triangle tables \citep{Regnier91completedetermination,FIKES1971189}, primarily to support plan modification and reuse. Triangle tables were later utilized as a preprocessing technique to take a sequential plan with conditional effects and extract partial ordering \citep{winner2002analyzing}. More recent work has applied partial weighted MaxSAT encodings to enhance execution flexibility by reducing the orderings in a plan \citep{maxsat}. Building on this, the \emph{action reinstantiation} extends the MaxSAT formulation with additional constraints, allowing operator parameter reassignment\citep{Waters_Nebel_Padgham_Sardina_2018,maxsat_reinst}.

The partial weighted MaxSAT problem extends the traditional satisfiability (SAT) problem by introducing two categories of clauses: \emph{hard} and \emph{soft}. Hard clauses function like those in the standard SAT problem and must always be satisfied. Soft clauses are not mandatory; instead, each is assigned a weight indicating its relative importance. 
Finding an assignment that satisfies all hard clauses and maximizes the cumulative weight of completely satisfied soft clauses is the goal.

\subsection{MR Encodings}
The task of achieving a plan's minimum reordering (MR) as a partial weighted MaxSAT problem instance is formulated by \cite{maxsat}. The solution of the instance corresponds to a POP, which they denote as the \emph{target} POP. For a given POP or sequential plan $\pi=\langle \mathcal{O}, \prec\rangle$, the encoding employs three categories of propositional variables$-$
\begin{itemize}
    \item $x_o$: Indicates the presence of operator $o$ in the \emph{target} POP for each $o \in \mathcal{O}$.  
    \item $\kappa (o_a, o_b)$: Indicates that the the \emph{target} POP includes the ordering constraint $o_a \prec o_b$ for each operator pair $o_a, o_b \in \mathcal{O}$.
    \item $\Upsilon (o_p, \langle x, d \rangle, o_c)$: Indicates that a causal link $o_p \xrightarrow{\langle x, d \rangle} o_c$ is present in the \emph{target} POP, for a pair of operator $o_p, o_c \in \mathcal{O}$, specifies that the operator $o_p$  produces $\langle x, d \rangle$ for the operator $o_c$, where $\langle x, d \rangle\in (prod_{o_p} \cap cons_{o_c})$.
\end{itemize}

The encoding procedure starts with defining hard clauses as Boolean formulae, which are then converted into conjunctive normal form (CNF), followed by the introduction of soft clauses and their associated weights. A soft clause with weight $k$ is represented by the syntax $\stackrel{k}{(\dots)}$, whereas no weight indication represents a hard clause. To guarantee correctness, the encoding defines formulae \eqref{eq:noselfloop} and \eqref{eq:transitive} that enforce acyclicity in the target POP, where \eqref{eq:transitive} ensures that the ordering constraints include the transitive closure. Formula \eqref{eq:initandgoal} includes the initial and goal operators. Operators are treated as universally quantified, and in the case of formula \eqref{eq:betweeninitandgoal}, it is assumed that $o_I \neq o_i \neq o_G$. Formula \eqref{eq:causallinkwithnothreat} guarantees that every causal link remains threat-free by enforcing the required promotion or demotion orderings, while formula \eqref{eq:causallinksupportedpre} ensures that each precondition is backed by an appropriate causal link.
\begin{align}
    & (\neg\kappa(o,o)) && \label{eq:noselfloop}\\
    & \kappa(o_p, o_q) \land \kappa(o_q, o_r) \xrightarrow{} \kappa(o_p, o_r) && \label{eq:transitive}\\
    & (x_{o_I}) \land (x_{o_G}) && \label{eq:initandgoal}\\
    & x_{o_i} \xrightarrow{} \kappa(o_I, o_i) \land \kappa(o_i, o_G) && \label{eq:betweeninitandgoal} \\
    & \Upsilon (o_p, \langle x, d \rangle, o_c) \xrightarrow{} \bigwedge\limits_{o_d: \langle x, d \rangle \in del_{o_d}} x_{o_d} \xrightarrow{} \kappa(o_d, o_p) \lor \kappa(o_c, o_d) && \label{eq:causallinkwithnothreat}\\
    & x_{o_c} \xrightarrow{} \bigwedge\limits_{\langle x, d \rangle\in cons_{o_c}}\ \bigvee\limits_{o_p:~\langle x, d \rangle\in prod_{o_p}} \kappa(o_p, o_c)\land \Upsilon(o_p, \langle x, d \rangle,o_c) && \label{eq:causallinksupportedpre}
\end{align}
For each operator and ordering variable in the encoding, a soft unit clause is introduced in formulae \eqref{eq:softclausefornegatingoperator} and \eqref{eq:softclausefornegatingordering} containing the negation of that variable. If any of these unit clauses is violated, it indicates that the operator or the ordering constraint associated with the variable of the violated clause is incorporated into the solution.
As a result, solutions with higher weights in the encoding correspond to target POPs that contain fewer ordering constraints.
\begin{align}
    & \stackrel{cost_o+|\mathcal{O}|^2+1}{(\neg x_o)}~~,~~ \forall o \in (\mathcal{O}\setminus\{o_I,o_G\}) && \label{eq:softclausefornegatingoperator}\\
    & \stackrel{1}{(\neg\kappa(o_i, o_j))}~~,~~ \forall o_i, o_j \in \mathcal{O} && \label{eq:softclausefornegatingordering}
\end{align}

\subsection{MRR Encodings}
The partial weighted MaxSAT encoding for achieving a plan’s minimum reordering is extended by \cite{maxsat_reinst} through allowing modifications to both operator parameters and ordering constraints to achieve greater flexibility, defined as minimum reinstantiated reordering (MRR). 
Variables are denoted by symbols like $x$, $y$, and $z$, constants by $c$, and terms by $t$, $u$, and $v$.
The definition of a term $t$ is an ordered series of items, where $t[i]$ is the $i$-th element in the sequence.
A mapping of terms with variables is designated as substitution $\Psi$; for instance, $\Psi = \{x_1/t_1,x_2/t_2, \dots , x_n/t_n\}$ maps $x_i$ to the corresponding term $t_i$ for all $i \in \{1,2,\dots,n\}$. The sets of variables and constants that appear in a given structure $\eta$ are denoted by $vars(\eta)$ and $consts(\eta)$. 
In this approach, a POP is represented as $\pi = \langle \mathcal{O}, \Psi, \prec \rangle$, where the set of operators is represented by $\mathcal{O}$, $\prec$ is a strict partial order over $\mathcal{O}$ that is transitively closed, and $\Psi$ is complete with respect to $\mathcal{O}$.

\begin{definition}

Let $\pi_p = \langle \mathcal{O}, \Psi, \prec \rangle$ and $\pi_q = \langle \mathcal{O}, \Psi', \prec' \rangle$ be two different POPs, then:
\begin{itemize}
    \item $\pi_q$ is a \textbf{reinstantiated reordering} of $\pi_p$ if $\pi_p$ and $\pi_q$ both are valid.
    \item $\pi_q$ is a \textbf{minimum reinstantiated reordering} of $\pi_p$ if $\pi_q$ is a reinstantiated reordering of $\pi_p$ and there does not exist a POP $\pi_r = \langle \mathcal{O}, \Psi'', \prec'' \rangle$ such that $\pi_r$ is a reinstantiated reordering of $\pi_p$ with $|\prec''| < |\prec'|$.
\end{itemize}
\end{definition}

An operator is defined as a tuple $o = \langle \mathit{vars}_o, \mathit{pre}_o, \mathit{eff}_o \rangle$, where $vars_o$ denotes the set of variables, and both $\mathit{eff}_o$ and $pre_o$ are the finite sets of facts containing $vars_o$'s variables. If both $pre_o$ and $\mathit{eff}_o$ are made up entirely of ground facts, then $o$ is considered as a ground operator. $\langle o_p, q(\overrightarrow{t}), o_c, q(\overrightarrow{u}) \rangle$ represents a causal link, where $q(\overrightarrow{u})$ and $q(\overrightarrow{t})$ are literals, with $q(\overrightarrow{t}) \in prod_o$ and $q(\overrightarrow{u}) \in cons_o$. The MRR encoding adds two additional categories of propositional variables in addition to the $x$ and $\kappa$ utilized in the MR encoding:
\begin{itemize}
    \item $\varepsilon (u,v)$: Encodes that in the final POP the condition $\Psi(t) = \Psi(u)$ must be satisfied.
    \item $\tau(o_p,\overrightarrow{t},o_c,\overrightarrow{u})$: Encodes that there exists some $q$ such that the $\langle o_p, q(\overrightarrow{t}), o_c, q(\overrightarrow{u}) \rangle$ causal link remains threat-free within the final POP.
\end{itemize}

To compute the MRR of a partial-order plan $\pi_p = \langle \mathcal{O}, \Psi, \prec \rangle$, the following encoding together with formulae \eqref{eq:noselfloop} through \eqref{eq:betweeninitandgoal} is used. Formulae \eqref{eq:9} and \eqref{eq:10} specify that the equality relation over variables and constants is both symmetric and transitive, whereas the mapping of each variable to a single object is guaranteed by Formula \eqref{eq:11}. Formulae \eqref{eq:12} and \eqref{eq:13} formalize the conditions required for the validity of the target POP.
\begin{align}
   & \varepsilon(u,v) \leftrightarrow \varepsilon(v,u) && \label{eq:9}\\
   & \varepsilon(u,v) \land \varepsilon(v,t) \rightarrow \varepsilon(u,t) && \label{eq:10}\\
   & \bigwedge\limits_{x\in vars_\mathcal{O}} (\bigvee\limits_{c\in consts_\mathcal{O}} \varepsilon(x,c)~~~\land 
   \bigwedge\limits_{\stackrel{c',c''\in consts_\mathcal{O}:} {c'\ne c''}} \neg\varepsilon(x,c') \lor \neg\varepsilon(x,c'')) && \label{eq:11}\\
    & \bigwedge\limits_{q(\overrightarrow{u})\in cons_{o_c}} \ \bigvee\limits_{q(\overrightarrow{t})\in prod_{o_p}} \tau(o_p, \overrightarrow{t}, o_c, \overrightarrow{u}) \land \kappa(o_p,o_c) && \label{eq:12}\\
    & \tau(o_p, \overrightarrow{t}, o_c, \overrightarrow{u}) \rightarrow  \bigwedge\limits_{1\leq i\leq|\overrightarrow{t}|} \varepsilon(\overrightarrow{t}[i],\overrightarrow{u}[i])~~~\land \bigwedge\limits_{\stackrel{q(\overrightarrow{v})\in del_{o_d}:}{\overrightarrow{t}= \overrightarrow{v}, o_d \ne o_c}} (\kappa (o_d, o_p) \lor \kappa(o_c, o_d)) && \label{eq:13}
\end{align}
If a POP has fewer than $k$ ordering constraints, determining if it admits a minimal reinstantiated reordering is $\mathit{NP}$\textit{-complete}.
It is impossible to approximate the problem of finding such a reordering within a constant factor. A key limitation of MRR is that substitutions are restricted to operators sharing the same name, but it does not permit replacing operator sets with different operator names or sizes.
\section{Explanation-based Order Generalization}
Another well-known approach, Explanation-based Order Generalization (EOG), employs validation structures as proofs of correctness and modifies plans to resolve inconsistencies \citep{Veloso2002,KAMBHAMPATI1994235,KAMBHAMPATI1992193}. EOG has been further refined to support conditional effects \citep{sabah_siddiqui_2022}.
The EOG deorders plans by creating a validation structure, creating a causal link for each operator’s preconditions, and using \textit{promotions} or \textit{demotions} to mitigate threats to those causal links \citep{kk,Veloso2002}. 
Given a total-order plan $\pi$ of a planning task $\Pi =\langle \mathcal{X}, \mathcal{O}, s_i, s_g \rangle$, EOG emulates the initial and goal conditions of $\Pi$ by augmenting $\pi$ with a couple of additional operators, $o_I$ and $o_G$. 
Specifically, $pre_{o_I}=\emptyset$, $\mathit{eff}_{o_I} = s_i$, $pre_{o_G} = s_g$, and $\mathit{eff}_{o_G}=\emptyset$, with $o_I \prec o_G$ and $o_I \prec o \prec o_G$ for all $o \in (\mathcal{O} \setminus \{o_I, o_G\})$. The validation structure is subsequently constructed, and threats are resolved by introducing promotion and demotion orderings. To minimize unnecessary transitive orderings, it systematically binds causal links to the earliest possible producers.
\subsection{Plan Deordering with Conditional Effects}
Conditional effects \citep{lipovetzky2019introduction} enable operators to present complex scenarios more effectively by allowing a single operator to produce different outcomes depending on the specific conditions. \cite{sabah_siddiqui_2022} introduce a method for annotating operator orderings in plans with conditional effects, allowing such plans to be transformed into a Partial-Order Plan (POP).

\begin{definition}
    Let $\Pi = \langle \mathcal{X}, \mathcal{O}_c, s_i, s_g \rangle$ be a planning task, where $\mathcal{O}_c$ is a set of a finite number of operators with conditional effects. An effect of an operator $o \in \mathcal{O}_c$ consists of triples $\langle cond, x, d \rangle$, forming a partial state where $cond$ denotes the effect condition, which may be empty. When $o$ is applicable in the state $s$, applying it yields the state $s' = apply(o, s)$, where $x \in \mathcal{X}$ takes the value $d \in D_x$ if and only if $cond \subseteq s$.
\end{definition}

\begin{definition}
    Let $\pi_{pop} = \langle \mathcal{O}, \prec \rangle$ be a POP and two operators $o_i, o_j \in \mathcal{O}$ with conditional effects.
    \begin{itemize}
        \item $o_i$ is considered as a \textbf{candidate producer} of $\langle x, d \rangle$ for operator $o_j$ iff:
        \begin{enumerate}
            \item $o_j \nprec o_i$,
            \item $\langle cond, x, d \rangle \in \mathit{eff}_{o_i}$ and there exists other candidate producer $o_p \in \mathcal{O}$ for every $\langle x', d' \rangle \in cond$ where $o_p \neq o_j$, and
            \item there exists no operator $o_k \in \mathcal{O}$ with $\langle cond_1, x, d_1 \rangle \in \mathit{eff}_{o_k}$ such that $o_j \nprec o_k \nprec o_i$, $d_1 \in (\mathcal{D}_{x} \setminus \{d\})$, and there exists other candidate producer $o_p \in \mathcal{O}$ for each $\langle x'', d'' \rangle \in cond_1$.
        \end{enumerate}
        \item $o_i$ is considered as an \textbf{earliest candidate producer} of $\langle x, d \rangle$ for operator $o_j$ iff $o_i$ is a candidate producer of $\langle x, d \rangle$ for $o_j$ and there does not exists any candidate producer $o_k$ of $\langle x, d \rangle$ for $o_j$ with $o_k \prec o_i$.
    \end{itemize}
\end{definition}

\begin{definition}
 Let $\pi_{pop}= \langle \mathcal{O, \prec \rangle}$ be a POP and $o \in \mathcal{O}$ be an operator with conditional effects. 
 The operator \textbf{o} produces, consumes, and deletes facts, which are represented by $\mathbf{prod_o}$, $\mathbf{cons_o}$, and $\mathbf{del_o}$, respectively. The fact$-$:
    \begin{itemize}
        \item $ \langle x, d \rangle\in cons_o$ if $\langle x, d \rangle\in pre_o$ or $\langle x, d \rangle \in cond$ such that $\langle cond,x',d' \rangle \in \mathit{eff}_o$ and there exists an operator ${o_k \in \mathcal{O}}$ with $o \xrightarrow{\langle x',d' \rangle} o_k$.

        \item $\langle x, d \rangle \in prod_o$ if $\langle cond,x, d \rangle \in \mathit{eff}_o$ and there exists other candidate producer $o_p \in \mathcal{O}$ for each $\langle x',d'\rangle \in cond$ such that $o_p \prec o$.
        \item $\langle x, d \rangle \in del_o$ if $x \notin vars(\mathit{cons}_o)$ or $cons_o(x) = d$, $\langle cond,x, d' \rangle \in \mathit{eff}_o$, where  $d' \in (\mathcal{D}_{x}\setminus \{d\})$, and there exists a candidate producer $o_p \in \mathcal{O}$ for each $\langle x',d''\rangle \in cond$.
    \end{itemize}
\end{definition}

$Re(o_a \prec o_b)$ denotes the ordering reasons between two operators $o_a$ and $o_b$ with conditional effects, which can fall into the categories PC, CT, TP, and OC.
$OC(\langle x, d\rangle)$ denotes an \emph{obstructor-consumer} relation for a fact $\langle x, d \rangle$, which occurs when $o_a$ prevents $o_b$ from receiving the fact $\langle x, d \rangle \in cond$, such that $\langle cond, x', d' \rangle \in \mathit{eff}_{o_b}$.
If there is an operator $o_k$ that yields $\langle x, d\rangle$ and $o_b \prec o_k \prec o_a$, then $OC(\langle x, d\rangle) \in Re(o_a \prec o_b)$ is considered \emph{threatened}.
This identified threat may be mitigated in two ways: through \emph{Promotion}, by including $pOC(\langle x, d\rangle)$ in $Re(o_k \prec o_a)$, or through \emph{Demotion}, by including $OCp(\langle x, d\rangle)$ in $Re(o_b \prec o_k)$.

The Deordering Plan with Conditional Effects (DConE) algorithm constructs causal links to build a validation framework for a POP. Similar to the EOG, DConE identifies causal links using the earliest candidate producers for each fact $\langle x, d\rangle$ required by an operator. While determining a candidate producer of $\langle x, d\rangle$ for an operator $o_i$, additional orderings may be needed due to $OC$ (obstructor-consumer) reasons. This occurs when an operator $o_d$ prevents another operator $o_p$ from providing $\langle x, d\rangle$ to $o_i$ through a conditional effect $\langle cond, x, d'\rangle \in \mathit{eff}_{o_d}$, where $d' \in (\mathcal{D}_x \setminus \{d\})$.
To address this, the algorithm introduces an ordering reason $OC(\langle x, d\rangle)$ in $Re(o_q \prec o_d)$ for some operator $o_q$ that obstructs the effect $\langle cond, x, d'\rangle$ of $o_d$ from being triggered. The ordering $o_q \prec o_d$ prevents $cond$ from holding, ensuring that $o_d$ no longer deletes $\langle x, d\rangle$. This process repeats iteratively until a valid candidate producer is found.
After establishing a causal link between two operators, DConE adds conditional causal links for each fact in the producer’s effect condition and resolves all threats to $PC$ and $OC$ reasons through promotion or demotion of the threatening operators.
\section{Block Deordering-based Approach}
Block deordering \citep{siddiqui_patrik_2012} identifies coherent sets of operators, known as blocks, to remove additional orderings from the plan.
Block deordering groups coherent operators into blocks to reduce the POP's ordering constraints, resulting in a block decomposed partial-order (BDPO) plan \citep{Siddiqui2015ContinuingPQ}. 
Each block contains a set of operators; operators from two disjoint blocks cannot interleave, allowing the blocks to be executed in any sequence.
Additionally, blocks can be nested, which allows a block to have one or more inner blocks. Overlapping between blocks is not permitted. A block fully enclosed within another is called an inner block, while a block not contained within any other is referred to as an outer block. Block deordering has also been leveraged to improve overall plan quality \citep{Siddiqui2015ContinuingPQ}, generate \emph{macro-actions} \citep{fazlul_Chrpa_2015}, and improve flexibility via block-substitution \citep{noor2024improving}. 
\begin{definition}
    A \textbf{BDPO plan} is denoted as $\pi_{bdpop} = \langle \mathcal{O}, \prec, \mathcal{B} \rangle$, where $\mathcal{O}$ be a collection of operators, $\prec$ is a collection of orderings within $\mathcal{O}$, and $\mathcal{B}$ is a set of blocks. For a block $b \in \mathcal{B}$, if $o_p, o_q \in b$ and $o_p \prec o_q$, then there does not exist an operator $o_r \notin b$ such that $o_p \prec o_r \prec o_q$. For any two blocks $b_p, b_q \in \mathcal{B}$, exactly one of the following holds: $b_p \subset b_q$, $b_q \subset b_p$, or $b_p \cap b_q = \emptyset$.
\end{definition}

A block b can be characterized by its preconditions and effects, similar to an operator. If some operator $o \in b$ consumes \atom{x}{d} and no operator $o' \in b$ provides \atom{x}{d} to operator $o$, then a fact \atom{x}{d} is one of the preconditions of $b$. Conversely, if an operator $o \in b$ generates \atom{x}{d} and no subsequent operator within $b$ modifies it, then \atom{x}{d} is included in the effects of $b$. Unlike an individual operator, a block can have multiple effects on the same or different variables. For example, if $o_i,o_j\in b$ with $o_i\nprec o_j$, $o_j\nprec o_i$, $\langle x, d \rangle \in \mathit{eff}_{o_i}$, and $\langle x, d' \rangle \in \mathit{eff}_{o_j}$, then a block $b$ has both $\langle x, d \rangle$ and $\langle x, d' \rangle$ as its effects (where $d \ne d'$).
The labels $PC$ (producer-consumer), $CT$ (consumer-threat), and $TP$ (threat-producer) can also be used to annotate ordering relations between blocks. 
\begin{definition}
    Let a BDPO plan $\pi_{bdpop} = \langle \mathcal{O}, \prec, \mathcal{B} \rangle$ with $b\in\mathcal{B}$ be a block, then:
    \begin{itemize}
        \item If $o_i \in b$ be an operator with $\langle x, d \rangle \in pre_{o_i}$ and there exists no operator $o_j \in b$ with $i\ne j$ that delivers $\langle x, d \rangle$ to $o_i$ via a causal link $o_j \xrightarrow{\langle x, d \rangle} o_i$, then the fact $\langle x, d \rangle \in pre_b$.
        \item If $o_i \in b$ be an operator with $\langle x, d \rangle \in \mathit{eff}_{o_i}$ and no subsequent operator $o_j \in b$ (with $o_i \prec o_j$) produces a different value $\langle x, d' \rangle$ for the same variable $x$ with $d' \in (\mathcal{D}_x \setminus \{d\})$, then the fact $\langle x, d \rangle \in \mathit{eff}_b$.
    \end{itemize}
\end{definition}

\begin{definition}
    For a block $b$, the sets of facts that it consumes, produces, and deletes are denoted as \textbf{cons\textsubscript{b}}, \textbf{prod\textsubscript{b}}, and \textbf{del\textsubscript{b}}, respectively. Where:
    \begin{itemize}
        \item A fact $\langle x, d \rangle \in cons_b$ if and only if $\langle x, d \rangle \in pre_b$.
        \item A fact $\langle x, d \rangle \in prod_b$ if and only if $\langle x, d \rangle \notin cons_b$, $\langle x, d \rangle \in \mathit{eff}_b$, and no other effect $\langle x, d' \rangle \in \mathit{eff}_b$ exists with $d' \in (\mathcal{D}_x \setminus \{d\})$.
        \item A fact $\langle x, d \rangle \in del_b$ if and only if either $x \notin vars(cons_b)$ or $cons_b(x) = d$, and there exists an effect $\langle x, d' \rangle \in \mathit{eff}_b$ with $d' \in (\mathcal{D}_x \setminus \{d\})$.
    \end{itemize}
\end{definition}

\begin{definition}
    Let a BDPO plan $\pi_{bdpop} = \langle \mathcal{O}, \prec, \mathcal{B} \rangle$ with a pair of blocks $b, b' \in \mathcal{B}$ and $\langle x, d \rangle \in cons_{b'}$, then:
    \begin{itemize}
        \item If $\langle x, d \rangle \in \mathit{eff}_{b}$, $b \prec b'$, and no other block $b'' \in \mathcal{B}$ exists such that $\langle x, d \rangle \in del_{b''}$ and $b' \nprec b'' \nprec b$, then block $b$ is a \textbf{candidate producer} of the fact $\langle x, d \rangle$ for $b'$.
        \item If there is no other candidate producer $b''$ of $\langle x, d \rangle$ for $b'$ with $b'' \prec b$, then block $b$ is considered as an \textbf{earliest candidate producer} of $\langle x, d \rangle$ for $b'$.
    \end{itemize}
\end{definition}


Block deordering builds a valid BDPO plan by taking a total-order plan as input.
The total-order plan is first converted into a POP $\pi = \langle \mathcal{O}, \prec \rangle$ using EOG.
Next, a block $b = \{o\}$ is created and added to $\mathcal{B}$ for every operator $o \in \mathcal{O}$ to construct an initial BDPO plan $\pi_{bdpop} = \langle \mathcal{O}, \prec, \mathcal{B} \rangle$.
Ordering constraint $b \prec b'$ is introduced for every ordering $o \prec o'$ present in $\prec$, where $o \in b$, $o' \in b'$, and $b, b' \in \mathcal{B}$.
The blocks with a single operator are denoted as \emph{primitive block} and the blocks with multiple operators are denoted as \emph{compound block}. The term \emph{block} is used in a general sense to cover both. Block deordering then applies a set of rules to remove additional orderings in $\pi_{bdpop}$. 

\begin{rules}
    \label{rule_1}
    A valid BDPO plan $\pi_{bdpop} = \langle \mathcal{O}, \prec, \mathcal{B} \rangle$ with an ordering $b_p \prec b_q$ and let $b$ be a block.
    \begin{enumerate}[label=\roman*.]
        \item \label{rule_1a} If $b_p \in b$, $b_q \notin b$, and $b_p \nprec b'$ for all $b' \in (b \setminus b_p)$, therefore, $PC(\langle x, d\rangle)$ can be eliminated from $Re(b_p \prec b_q)$ if $\langle x, d\rangle \in pre_b$ and there exists $b_p \notin b$ which can create causal links $b_p \xrightarrow{\langle x, d\rangle} b_q$ and $b_p \xrightarrow{\langle x, d\rangle} b$.
        \item \label{rule_1b} If $b_p \in b$, $b_q \notin b$, and $b \cap b_q = \emptyset$, therefore, $CT(\langle x, d\rangle)$ can be eliminated from $Re(b_p \prec b_q)$ if $\langle x, d\rangle \notin cons_b$. 
        \item \label{rule_1c} If $b_p \notin b$, $b_q \in b$, and $b_p \cap b = \emptyset$, therefore, $CT(\langle x, d\rangle)$ can be eliminated from $Re(b_p \prec b_q)$ if $\langle x, d\rangle \notin del_b$.    
        \item\label{rule_1d} If $b_q \in b$ and $b_p \notin b$, therefore, $TP(\langle x, d\rangle)$ can be eliminated from $Re(b_p \prec b_q)$ if $b$ contains all blocks $b'$ with $b_q \xrightarrow{\langle x, d\rangle} b'$.
    \end{enumerate}
\end{rules}

Block deordering begins by examining each ordering in the initial BDPO plan from top to bottom, attempting to remove them greedily. 
If Rule \ref{rule_1} can eliminate all of the ordering reasons for an ordering $b_p \prec b_q$, then that ordering is eliminated.
The ordering $b_p \prec b_q$ is kept, and the algorithm proceeds to the subsequent ordering if some reasons cannot be eliminated.
Whenever an ordering is successfully removed, the modified BDPO plan is returned by the algorithm. The deordering process is then restarted by the algorithm from the top of this most recent plan. Until no more orderings can be eliminated from the most recent BDPO plan, this iterative process keeps going.

\subsection{Flexibility Improvement via Block-Substitution}
Block-substitution \citep{noor2024improving} permits changing a block inside a valid BDPO plan without compromising the plan's validity. 
The block that is being replaced is referred to as the \emph{original block}, and the block that is replacing it is referred to as the \emph{substituting block}.
Block substitution permits the replacement block to originate either from within the plan or from an external source. When the replacement block is drawn from within the same plan, the process is referred to as an internal block substitution. 
Establishing causal links for the replacing block's preconditions is required for the substitution process, as well as restoring all causal links that the original block had previously supported. Furthermore, any threats introduced by the substitution must be identified and resolved to maintain the validity of the plan.
\begin{definition}
    Let $\pi_{bdpop}=\langle \mathcal{O}, \prec, \mathcal{B} \rangle$ be a valid BDPO plan, $b\in \mathcal{B}$, and a subplan $\tilde{b}=\langle \tilde{\mathcal{O}},\tilde{\prec}\rangle$ with $\tilde{\mathcal{O}} \subset \mathcal{O}$. A new BDPO plan $\pi'_{bdpop}= \langle \mathcal{O'}, \prec', \mathcal{B'} \rangle$ is produced if $b$ is substituted with $\tilde{b}$ where $b\notin \mathcal{B}'$ and $\tilde{b}\in \mathcal{B}'$. If $\pi'_{bdpop}$ is a valid BDPO plan, then the block-substitution is considered \emph{valid}.
\end{definition}
\begin{definition} \label{resolve}
    Let $\pi_{bdpop}=\langle \mathcal{O}, \prec, \mathcal{B} \rangle$ be a valid BDPO plan and some causal link $b_p \xrightarrow{\langle x, d \rangle} b_c$ is threatened by $b_t \in \mathcal{B}$, where $\langle x, d \rangle \in del_{b_t}$ and $b_p, b_c \in \mathcal{B}$.
    A threat is considered resolved if a \textbf{threat-resolution strategy} from the following can be used without creating a cycle in $\pi_{bdpop}$:
    \begin{enumerate}[label=\roman*.]
        \item \textit{Promotion}: introduce the ordering $b_t \prec b_p$ into $\prec$.
        \item \textit{Demotion}: introduce the ordering $b_c \prec b_t$ into $\prec$.
        \item \textit{Internal substitution}: replace $b_t$ with $b_c$, or $b_c$ with $b_t$.
    \end{enumerate}
\end{definition}
The procedure replaces a block $b$ with a substituting block $\tilde{b}$.  
The process starts by making sure that every precondition of the $\tilde{b}$ is supported by a causal link, but $\tilde{b}$'s preconditions are already satisfied if it is internal. 
For an external block, $\tilde{b}$ is first included in the plan, and causal links are established using the \emph{earliest candidate producers} for each precondition; if any precondition cannot be supported, the substitution fails. 
Then, using $\tilde{b}$ as the new producer, every causal link that block $b$ initially supported is restored.
If $\tilde{b}$ does not produce a required fact for any dependent block, the substitution is unsuccessful. Once these causal links are in place, $b$ is removed from the plan. 
Finally, any threats introduced by the substitution are identified and resolved using threat-resolution strategies, 
including promotion, demotion, or internal block-substitution. In some situations, replacing a conflicting block 
with $\tilde{b}$ resolves threats while maintaining plan validity.
The block-substitution process has an overall worst-case complexity of $O(n^2 p^2)$, where $p$ is the highest possible number of facts in any operator's effect or precondition and $n$ is the number of operators in the plan.

Flexibility Improvement via Block-Substitution (FIBS) \citep{noor2024improving} algorithm generates a valid BDPO plan given a valid total-order plan $\pi$ as input.
First, using EOG, a partial-order plan $\pi_{pop} = (\mathcal{O}, \prec)$ is created from $\pi$. 
A BDPO plan $\pi_{bdpop} = \langle \mathcal{O}, \prec, \mathcal{B} \rangle$ is then created from this partial-order plan by constructing a block $b = \{o\}$ for each $o \in \mathcal{O}$. 
In the next phase, ordering constraints are removed from $\pi_{bdpop}$ by substituting blocks, where initially only primitive blocks are considered, since no compound blocks have yet been formed. 
Subsequently, block deordering introduces compound blocks to eliminate additional orderings, further increasing plan flexibility. 
Finally, the procedure of substituting blocks is employed once more to replace both compound and primitive blocks, minimizing orderings in the plan. 
The procedure of substituting blocks examines each basic ordering $b_i \prec b_j$ in the BDPO plan and attempts to remove it by substituting $b_j$ or, if necessary, $b_i$, while ensuring causal links and plan validity are maintained. 
After a successful removal, the procedure is restarted from the current BDPO plan's beginning, iterating until no further orderings can be eliminated. 
This two-stage application, before and after block deordering, allows the algorithm to separately evaluate the impact of primitive and compound block substitutions on overall plan flexibility.


FIBS significantly enhances plan flexibility while reducing computational and plan costs through efficient block-substitution within BDPO plans, outperforming existing approaches like EOG and MaxSAT reorderings. However, FIBS restricts candidate subplans for substitution to blocks within a BDPO plan, leaving other potentially useful subplans unexplored, and it has not yet been evaluated across diverse planning paradigms or with planners beyond LAMA.

\subsection{Concurrency Improvement via Block-
Substitution}

Partial-order plans enable parallel execution by indicating which operators cannot run concurrently. A parallel plan is a POP that permits concurrent operator execution. 
\cite{noor2025improving} transform sequential plans into parallel ones and define non-concurrency constraints in FDR. 
They improved the flexibility of a plan's execution by expanding block deordering and block substitution approaches, thereby exploiting potential parallelism to minimize overall execution time.

\begin{definition}
A \textbf{parallel plan} is denoted as $\pi_\mathcal{P}=\langle \mathcal{O}, \prec, \#\rangle$, which basically extends a POP $\langle \mathcal{O}, \prec\rangle$ with an irreflexive and symmetric relation $\#$, which represents non-concurrency constraints over $\mathcal{O}$.
Two operators, $o_a$ and $o_b$, have a non-concurrency constraint, expressed as $o_a\# o_b$, if $cons_{o_a}(x) \ne cons_{o_b}(x)$, $prod_{o_a}(x) \ne prod_{o_b}(x)$, or $cons_{o_a}(x) \ne prod_{o_b}(x)$ for any variable $x\in \mathcal{X}$, which means that they cannot be executed simultaneously.
\end{definition}

\begin{definition}
    Let $\pi_\mathcal{P}=\langle \mathcal{O}, \prec, \#\rangle$ be a parallel plan. The ratio of the operator pairs that can be executed simultaneously to the maximum possible operator pairs is the \textbf{concurrent flexibility} value of the parallel plan $\pi_{\mathcal{P}}$, as specified in \eqref{eq:cflex}, expressed as $\mathit{cflex}(\pi_{\mathcal{P}})$.
    \begin{equation}
        \mathit{cflex}(\pi_{\mathcal{P}}) = 1- \frac{ \sum\limits_{1\le b<a\le |\mathcal{O}|}
        \begin{cases}
        1 & \text{if $o_a \prec o_b$, $o_b \prec o_a$, or $o_a\# o_b$}\\
        0 & \text{otherwise}
        \end{cases}
        }{\sum_{n=1}^{|\mathcal{O}|-1}n}
        \label{eq:cflex}
    \end{equation}
\end{definition}

Two unordered operators of a partial-order plan can be executed in any sequence without invalidating the plan, but they can’t always run at the same time because they might use the same resources. To safely run operators in parallel, we must make sure they don’t interfere with each other. 
\cite{Knoblock1994} grouped parallel plans based on how operators depend on each other, describing the simplest kind as those where operators are independent with respect to the goal, meaning the final result is the same whether they run one after another or together.
\cite{noor2025improving} focus on using such independent operators to speed up plan execution, not because parallelism is required, but because it can make plans more efficient. 
Methods like \cite{Knoblock1994} and \cite{Boutilier_2001} handle this by explicitly defining which resources each operator used, but none established how to identify parallelism without such information.
\cite{noor2025improving} extended the idea of the BDPO plan by adding the non-concurrency relation to allow blocks to be executed in parallel.

\begin{definition}
    A \textbf{parallel block decomposed partial-order} (PBDPO) plan is denoted as $\pi_{pbdp}=\langle \mathcal{O}, \prec, \mathcal{B}, \#\rangle$, which basically extends a BDPO plan $\langle \mathcal{O}, \prec, \mathcal{B}\rangle$ with an symmetric, irreflexive relation $\#$ over $\mathcal{B}$, depicting non-concurrency constraints between blocks. 
    Blocks $b_p$ and $b_q$ are considered non-concurrent, expressed as $b_p \# b_q$, if two primitive operators $o_p \in b_p$ and $o_q \in b_q$ are non-concurrent ($o_p \# o_q$). Variables causing this non-concurrency are denoted by the set $vars(b_p \# b_q)$.
\end{definition}

\begin{definition}
    Let $\pi_{pbdp}=\langle \mathcal{O}, \prec, \mathcal{B}, \#\rangle$ be a valid parallel BDPO plan with $b_p\# b_q$, where blocks $b_p, b_q\in \mathcal{B}$ are two disjoint blocks. The $b_p\# b_q$ is considered \textbf{necessary non-concurrency constraint} within $\pi_{pbdp}$ if$-$
    \begin{enumerate}[label=\roman*.]
        \item $b_p\nprec b_q\nprec b_p$, and
        \item if $b_p$ is in the block $b_x \in B$ , then $b_q$ is also in the block $b_x$ .
    \end{enumerate}
\end{definition}
The Concurrent Flexibility Improvement via Block-Substitution (CIBS) algorithm generates a valid PBDPO plan from a valid total-order plan $\pi$ for a planning task $\Pi$.
The CIBS algorithm extends blocks before substitution. 
It selects two blocks $b_p$ and $b_q$ with $b_p\#b_q$ in a PBDPO plan $\pi_{pbdp}$ with respect to a planning task $\Pi$. 
Then, the procedure includes the necessary blocks to the block $b_p$ to achieve a new block $b_p'$ so that another block $\hat{b}_p$ can be used in place of $b_p'$ where $vars(\hat{b}_p\#b_q)=\emptyset$.

The CIBS algorithm enhances the concurrent flexibility through block-substitution in three steps.
In the first step, EOG is applied to convert $\pi$ into a parallel plan $\pi_{eog}= \langle \mathcal{O}, \prec,\#\rangle$ first, and then non-concurrency constraints between every pair of operators are identified.
In the next step, cohesive operators are encapsulated into blocks using block deordering, which removes ordering constraints from $\pi_{eog}$ and yields a PBDPO plan $\pi_{pbdp} = \langle \mathcal{O}, \prec, \mathcal{B}, \#\rangle$.
In the last stage, blocks are replaced to improve the concurrent flexibility of $\pi_{pbdp}$.

The CIBS effectively enhances concurrent flexibility across a wide range of planning domains, outperforming both EOG and block deordering in most cases. Its consistent improvements in $\mathit{cflex}$, particularly in domains rich with resource-based interactions, show the benefit of incorporating substitution alongside block deordering. The observed correlations suggest that CIBS scales reasonably well with plan complexity, though execution time grows notably with larger plan sizes and more variables. However, the performance drop in extensive plans and the limited impact of block deordering on $\mathit{cflex}$ highlight scalability and efficiency challenges. Additionally, the algorithm’s reliance on available resources for substitution restricts its applicability in domains lacking such elements, indicating room for improvement in generalizing CIBS to more diverse planning tasks.
\section{Analysis}
\label{analysis}

The optimization capabilities of the six algorithms are summarized in Table~\ref{tab:capability} across six categories: ordering, action handling, parameter handling, plan structure, concurrency, and complexity and optimality. 

\begin{table*}[htbp]
\centering
\caption{Optimization capability comparison among algorithms}
\small
\begin{tabular}{@{}lc@{\hspace{8pt}}c@{\hspace{8pt}}c@{\hspace{8pt}}c@{\hspace{8pt}}c@{\hspace{8pt}}c@{}}
\toprule
\textbf{Criteria} & \textbf{EOG} & \textbf{BD} & \textbf{MR} & \textbf{MRR} & \textbf{FIBS} & \textbf{CIBS} \\
\midrule

\multicolumn{7}{l}{\textbf{Ordering}} \\
\quad Remove Orderings & Yes & Yes & Yes & Yes & Yes & Yes \\
\quad Arbitrary Reordering & No & No & Yes & Yes & No & No \\
\quad Block-level Deordering & No & Yes & No & No & Yes & Yes \\
\quad Minimizes Orderings & No & No & Yes & Yes & No & No \\

\midrule
\multicolumn{7}{l}{\textbf{Action Handling}} \\
\quad Remove Redundant Actions & No & No & Yes & Yes & No & No \\
\quad Introduce New Actions & No & No & No & No & Yes & Yes \\
\quad Replace Action Sets (Subplans) & No & No & No & No & Yes & Yes \\
\quad Replace Actions of Different Name & No & No & No & No & Yes & Yes \\

\midrule
\multicolumn{7}{l}{\textbf{Parameter Handling}} \\
\quad Rebind Parameters & No & No & No & Yes & No & No \\
\quad Same-name Operator Rebinding Only & - & - & - & Yes & - & - \\

\midrule
\multicolumn{7}{l}{\textbf{Plan structure}} \\
\quad Produces Partial-order Plan & Yes & Yes & Yes & Yes & Yes & Yes \\
\quad Produces BDPO Plan & No & Yes & No & No & Yes & Yes \\
\quad Produces Parallel Plan & No & No & No & No & No & Yes \\
\quad Produces Parallel BDPO Plan & No & No & No & No & No & Yes \\
\quad External Planner for Subplans & No & No & No & No & Yes & Yes \\

\midrule
\multicolumn{7}{l}{\textbf{Concurrency}} \\
\quad Non-concurrency Constraints (\textbf{\#}) & No & No & No & No & No & Yes \\
\quad Minimizes \textbf{\#} & No & No & No & No & No & Yes \\
\quad Improves Parallel Execution & No & Partial & No & No & Partial & Yes \\
\quad Uses FDR for \textbf{\#} Conditions & No & No & No & No & No & Yes \\
\quad Block Extension via DTG Analysis & No & No & No & No & No & Yes \\

\midrule
\multicolumn{7}{l}{\textbf{Complexity and Optimality}} \\
\quad Polynomial-time Algorithm & Yes & Yes & No & No & Partial & Partial \\
\quad MaxSAT-based Encoding & No & No & Yes & Yes & No & No \\
\quad Anytime / Iterative Improvement & No & No & No & No & Yes & Yes \\
\quad Composable on top of MR/MRR & No & No & - & - & Yes & No \\

\bottomrule
\end{tabular}
\label{tab:capability}
\end{table*}

All six algorithms remove unnecessary orderings from a plan, but they differ in scope. EOG operates at the operator level and produces a standard POP without block structure. Block Deordering (BD) groups coherent operators into blocks and eliminates more orderings than EOG by treating each block as a unit. MR and MRR allow arbitrary reordering and find minimum reorderings via MaxSAT encodings, though neither produces block-structured plans. Flexibility Improvement via Block-Substitution (FIBS) and Concurrency Improvement via Block-Substitution (CIBS) use block-level deordering but, unlike MR and MRR, do not seek a globally minimum reordering; instead, they improve the plan iteratively through substitution.

On action handling, EOG and BD leave the operator set unchanged. MR removes redundant actions as part of its encoding. MRR does the same and additionally reorders operator parameters, though only within operators of the same name. FIBS and CIBS replace entire subplans with new ones drawn from the planning task, including operators with different names and different numbers of actions. Block-substitution is therefore the only mechanism in the table that can introduce actions absent from the original plan.

Parameter rebinding is exclusive to MRR, and even there it is constrained to same-name operators. All other algorithms treat operator parameters as fixed.

All six algorithms produce a valid POP. BD, FIBS, and CIBS additionally produce BDPO plans by organizing operators into blocks. Only CIBS produces a parallel plan by incorporating a non-concurrency constraint relation ($\#$), which is a parallel block decomposed (PBD) plan. FIBS and CIBS both use an external planner to generate candidate subplans for substitution; the remaining algorithms are self-contained.

The concurrency category distinguishes CIBS from everything else in the table. EOG, BD, MR, MRR, and FIBS do not model non-concurrency constraints ($\#$). CIBS formalizes necessary and sufficient conditions for non-concurrency using FDR variables, uses a Domain Transition Graph (DTG) to determine when a block must be extended before substitution, and directly minimizes $\#$ constraints to maximize parallel execution flexibility.

EOG and BD run in polynomial time. MR and MRR are NP-complete due to their MaxSAT formulations. FIBS and CIBS are partially polynomial: the block-substitution procedure runs in $O(n^2p^2)$ time, but both use an external planner, which is not guaranteed to be polynomial. Both are anytime algorithms that produce a valid result at any point during execution. Finally, FIBS can be applied on top of MR or MRR-generated POPs for further flexibility gains; CIBS takes a sequential plan directly as input and is not designed to compose with MaxSAT reorderings.
\section{Evaluation}
We evaluate seven plan execution flexibility-improving methods on 3,345 plans from 46 IPC domains. These methods are EOG, BD, MR, MRR, FIBS, FIBS applied on MR output (FIBS+MR), and FIBS applied on MRR output (FIBS+MRR). EOG and BD achieve full coverage by construction, while MR, MRR, FIBS+MR, and FIBS+MRR solve 81.4\%, 47.8\%, 48.2\%, and 21.0\% of plans, respectively. When an algorithm fails to find a solution, we assume its flex value with the EOG flex for that plan, which is the lowest flex any plan achieves before any reordering or substitution.

We measure flexibility using the flex metric defined in \eqref{eq:flex}. For a multi-dimensional comparison, we compute five normalized scores per algorithm: Flex (mean flex value), Coverage (proportion of plans solved), Speed (inverse mean execution time), Consistency (inverse coefficient of variation of flex), and Unique Best (proportion of plans on which the algorithm achieves the strictly highest flex).
All five are min-max normalized, so a value of 1.0 denotes the best algorithm on that dimension.

\subsection{Flex-Time Trade-off}
\begin{figure}[htbp]
    \centering
    \includegraphics[width=\textwidth]{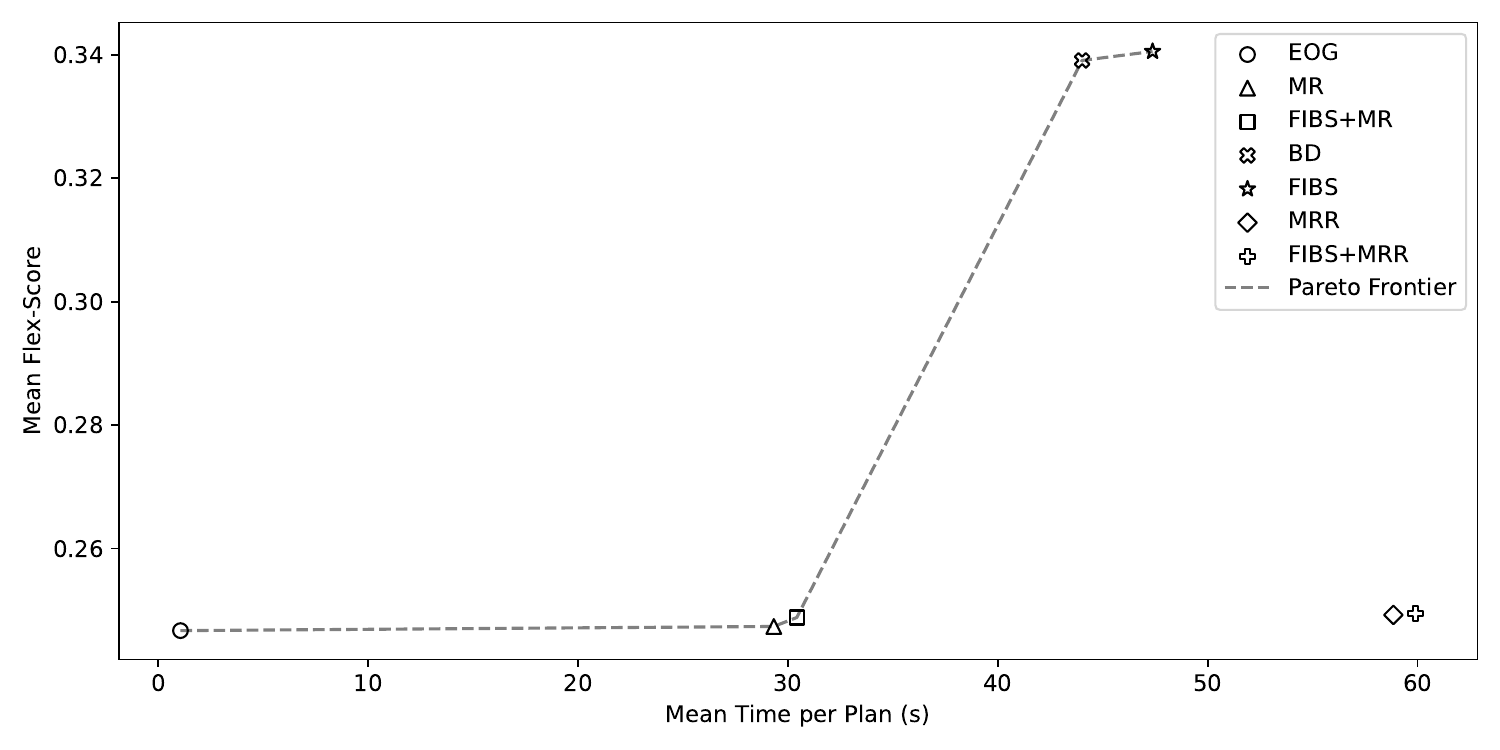}
    \caption{Mean flex score vs. mean planning time with Pareto Frontier.}
    \label{fig:pareto}
\end{figure}

Figure~\ref{fig:pareto} plots mean flex against mean time per plan for all seven methods, with the Pareto frontier connecting those not dominated on both dimensions at once. EOG is the fastest at 1.06 seconds with a mean flex of 0.247. MR and FIBS+MR reach mean flex values of 0.247 and 0.249 at 29.3 and 30.4 seconds, respectively, both on the frontier. BD and FIBS improve mean flex to 0.339 and 0.341 at 44.0 and 47.4 seconds, also on the frontier. MRR and FIBS+MRR take 58.8 and 59.9 seconds but reach only 0.249 mean flex each, placing them off the frontier. The Pareto front therefore has two natural operating points: fast with modest gains (EOG, MR) and moderate cost with substantially better flex (BD, FIBS). MRR and FIBS+MRR roughly double MR's computation time without improving its position on the frontier.

\subsection{Flex Efficiency}
\begin{figure}[htbp]
    \centering
    \includegraphics[width=\textwidth]{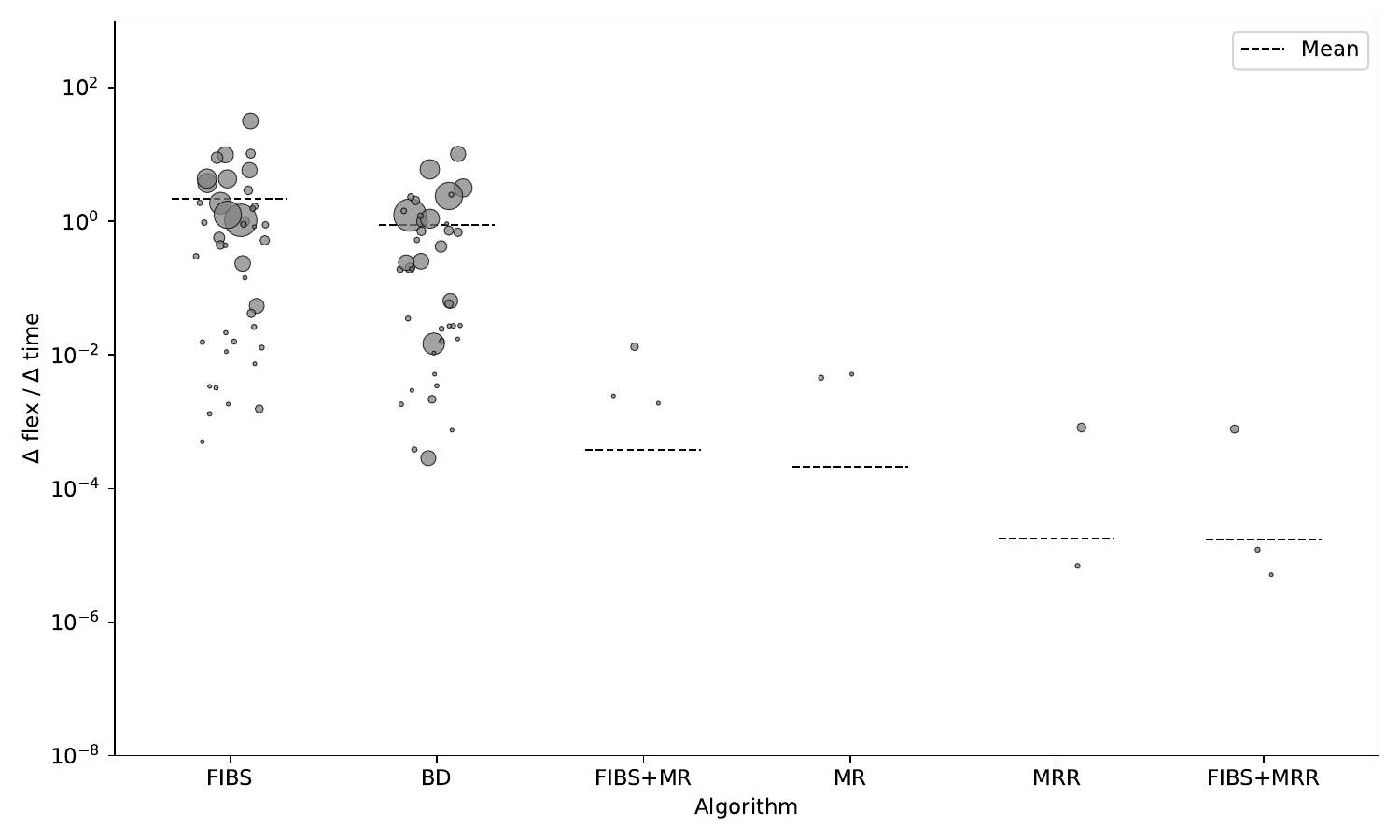}
    \caption{Flex efficiency}
    \label{fig:flex_efficiency}
\end{figure}

Figure~\ref{fig:flex_efficiency} shows flex efficiency per algorithm and domain, measured as flex gain over EOG per additional second spent ($\Delta$flex/$\Delta$time). 
BD has the highest median efficiency across domains: block deordering removes a large number of ordering constraints at relatively low per-plan overhead. 
FIBS reaches comparable efficiency because its substitution phases build directly on BD. 
MR and FIBS+MR cluster at intermediate values. 
MRR and FIBS+MRR fall to the lowest efficiency, as MRR's encoding cost is large relative to the flex improvement it achieves. 

Bubble size in Figure~\ref{fig:flex_efficiency} encodes absolute mean flex gain per domain; the largest bubbles belong to BD and FIBS, confirming that these two algorithms also produce the greatest absolute improvements.

\subsection{Statistical Effect Size}

\begin{figure}[htbp]
    \centering
    \includegraphics[width=\textwidth]{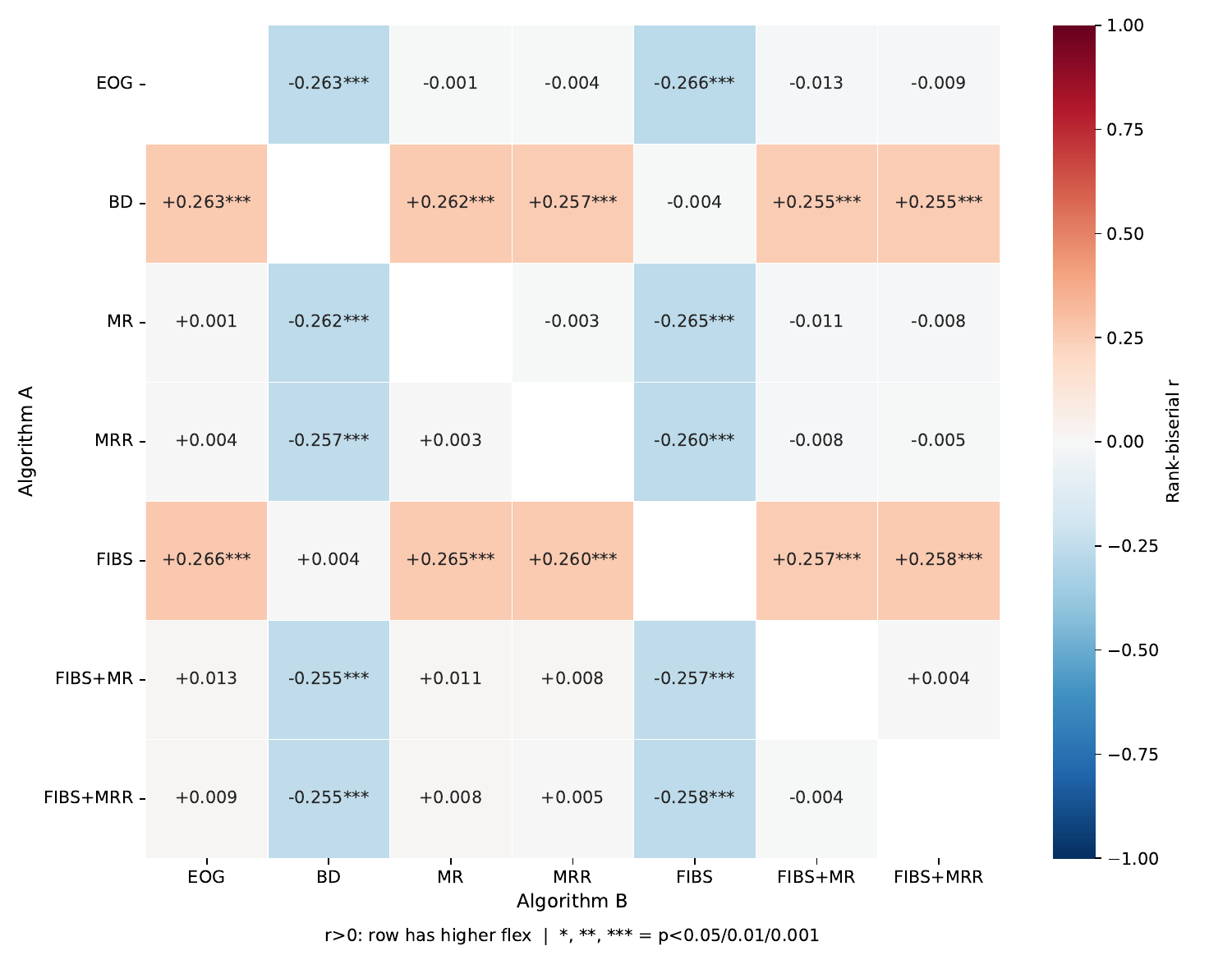}
    \caption{Effect Size - Rank-biserial r}
    \label{fig:effect_size_heatmap}
\end{figure}

Figure~\ref{fig:effect_size_heatmap} presents a pairwise effect-size heatmap using the rank-biserial correlation $r$ from a two-sided Mann–Whitney U test on the flex distributions. A positive $r$ in row $A$ and column $B$ means algorithm $A$ produces higher flex on more plans than algorithm $B$; significance levels are marked as *($p<0.05$), ** ($p<0.01$), and *** ($p<0.001$).

BD and FIBS show strong positive $r$ values against all other algorithms, with statistically significant differences in the majority of pairwise comparisons. The effect size between BD and FIBS itself is small and non-significant, so the two algorithms perform equivalently at the population level despite FIBS having a marginally higher mean. EOG, MR, MRR, FIBS+MR, and FIBS+MRR cluster together, with near-zero or weakly negative $r$ values against the block-based methods. Within this cluster, pairwise effect sizes are consistently small and non-significant, indicating that the MaxSAT-based methods do not meaningfully separate from one another in terms of flex on the full plan distribution.

\subsection{Multi-Criteria Comparison}

\begin{figure}[htbp]
    \centering
    \includegraphics[width=\linewidth]{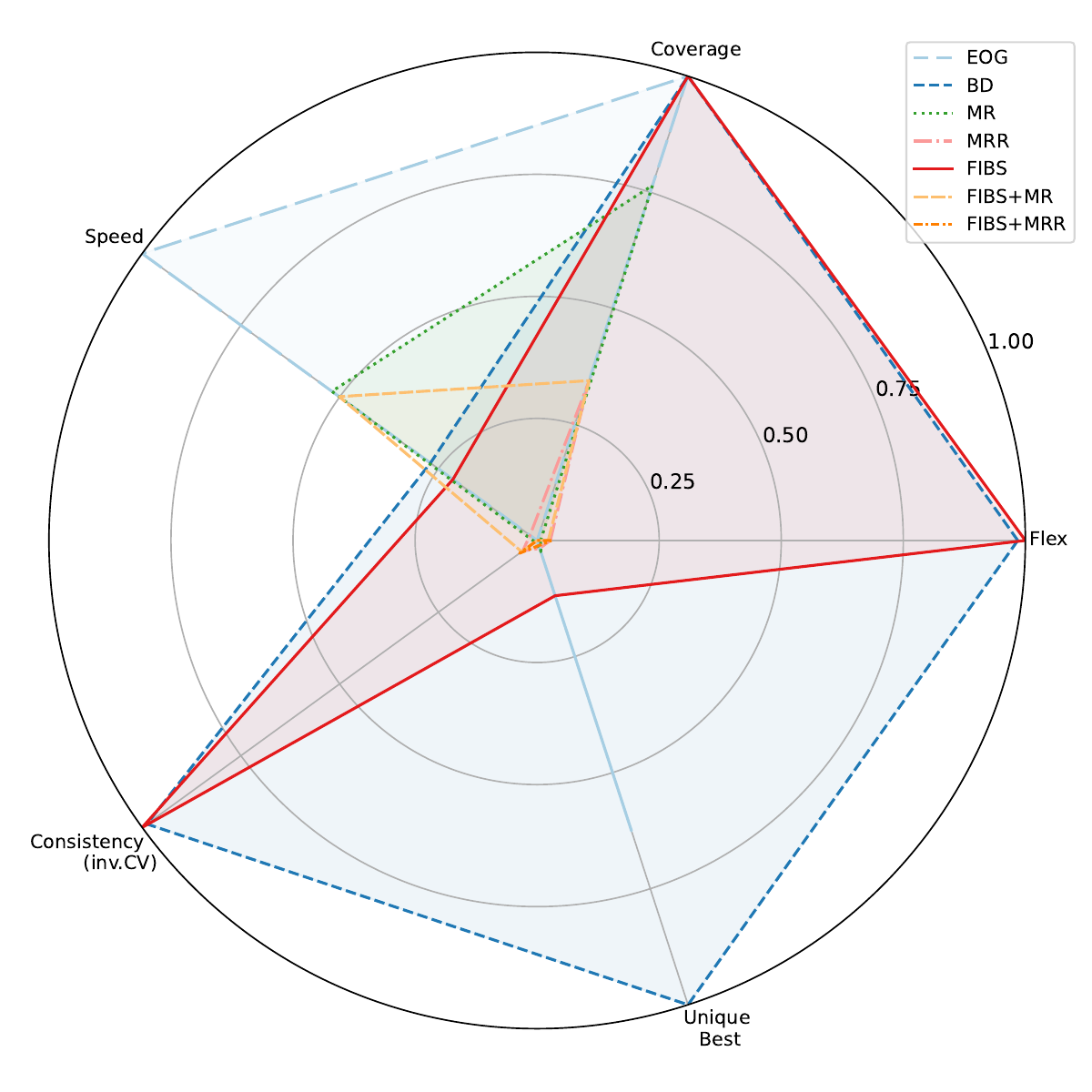}
    \caption{Radar Chart}
    \label{fig:radar_chart}
\end{figure}

Figure~\ref{fig:radar_chart} places all seven methods on a radar chart with one axis per normalized dimension; the outer rim corresponds to the best score on each axis.
BD and FIBS occupy the largest area, with normalized Flex scores of 0.985 and 1.0, perfect Coverage, and perfect Consistency. FIBS has a Unique Best score of 0.119, meaning it produces the strictly highest flex on roughly 12\% of individual plans. BD's normalized Unique Best score is 1.0, as it achieves the best flex on the largest number of plans overall. EOG scores perfectly on Speed and Coverage, and its normalized Unique Best of 0.625 is notably high: since EOG is used as the fallback for plans where MaxSAT methods time out, it retains the best score wherever all alternatives fail to terminate.

MR achieves moderate Speed (0.520) and Coverage (0.765) but a near-zero normalized Flex score relative to BD and FIBS. MRR scores near zero on Speed, Coverage, and Unique Best. FIBS+MR and FIBS+MRR also score near zero on Unique Best, since when the MaxSAT component does find a solution, FIBS adds only a small additional flex increment, and the combined pipeline rarely produces the globally best result for any individual plan.

\section{Conclusion}
\label{conclusion}

The methods reviewed in this study approach plan flexibility from three distinct angles: causal-link reasoning (POCL, EOG), operator coherence and block structure (BD, FIBS, CIBS), and constraint optimization (MR, MRR). Each angle captures something real about what makes a plan flexible, but the empirical results show that these angles are not equally productive. Restructuring causal dependencies at the block level, without any guarantee of minimality, yields substantially more flexibility than minimizing ordering constraints within a fixed causal structure. This is the central finding of the study, and is somewhat counterintuitive given how much theoretical work has gone into minimum reordering.

The theoretical promise of finding minimum reorderings with the MaxSAT-based methods is legitimate. Yet minimum reordering within a fixed action set is a ceiling set by the original plan's causal structure. Block deordering raises that ceiling by reorganizing how causal dependencies are grouped, which exposes ordering constraints that would otherwise appear necessary. Block-substitution raises it further by replacing subplans with alternatives that carry different causal entailments entirely. BD's greedy, polynomial-time heuristic therefore outperforms MR's NP-complete search on a broad collection of IPC benchmarks.

Minimum reordering has been studied extensively as an optimization target, but the quality of the causal structure itself has received less attention. The success of block-substitution suggests that reordering and action-set modification interact in non-trivial ways. MRR takes a step in this direction by allowing parameter rebinding, but only within same-name operators, and the evaluation shows this restriction is too tight to close the gap with BD. A method that jointly optimizes ordering and action selection without the coverage failures of MaxSAT encodings is still missing.

CIBS introduces a separate question that none of the other methods address. Two unordered operators may still be concurrent if they do not share a state variable. CIBS introduces $\mathit{cflex}$ to account for this, instead of relying on $\mathit{flex}$, and the gap between the two metrics is non-trivial in resource-rich domains. However, whether a domain-independent characterization of practical concurrency is achievable, and how to extend concurrent flexibility improvement to domains without explicit resource contention, remain open.

\section*{Statements and Declarations}
\subsection*{Competing Interests}
The authors have no relevant financial or non-financial interests to disclose.

\bibliography{sn-bibliography}

@article{noor2024improving,
  title={Improving Plan Execution Flexibility using Block-Substitution},
  author={Noor, Sabah Binte and Siddiqui, Fazlul Hasan},
  journal={arXiv preprint arXiv:2406.03091},
  year={2024}
}

@InProceedings{sabah_siddiqui_2022,
author="Noor, Sabah Binte
and Siddiqui, Fazlul Hasan",
editor="Arai, Kohei",
title="{P}lan Deordering with Conditional Effects" ,
booktitle="Intelligent Systems and Applications" ,
year="2022",
publisher="Springer International Publishing" ,
pages="852--870",
isbn="978-3-031-16078-3"
}

@article{noor2025improving,
  title={Improving execution concurrency in partial-order plans via block-substitution},
  author={Noor, Sabah Binte and Siddiqui, Fazlul Hasan},
  journal={Autonomous Agents and Multi-Agent Systems},
  volume={39},
  number={1},
  pages={1--43},
  year={2025},
  publisher={Springer}
}

@Inproceedings{siddiqui_patrik_2012,
    author="Siddiqui, Fazlul Hasan
    and Haslum, Patrik",
    editor="Thielscher, Michael
    and Zhang, Dongmo",
    title="Block-Structured Plan Deordering",
    booktitle="AI 2012: Advances in Artificial Intelligence",
    year="2012",
    publisher="Springer Berlin Heidelberg",
    address="Berlin, Heidelberg",
    pages="803--814",
}

@inproceedings{fazlul_Chrpa_2015,
author = {Chrpa, Luk\'{a}\v{s} and Siddiqui, Fazlul Hasan},
title = {{E}xploiting Block Deordering for Improving Planners Efficiency},
year = {2015},
isbn = {9781577357384},
publisher = {AAAI Press},
booktitle = {Proceedings of the 24th International Conference on Artificial Intelligence},
pages = {1537–1543},
numpages = {7},
location = {Buenos Aires, Argentina},
series = {IJCAI'15}
}

@article{Siddiqui2015ContinuingPQ,
  title={{C}ontinuing Plan Quality Optimisation},
  author={Fazlul Hasan Siddiqui and Patrik Haslum},
  journal = {Journal of Artificial Intelligence Research},
  year={2015},
  volume={54},
  pages={369-435}
}

@inproceedings{Anderson1988,
    author = {Anderson, John S. and Farley, Arthur M.},
    title = {{P}lan Abstraction Based on Operator Generalization},
    year = {1988},
    publisher = {AAAI Press},
    booktitle = {Proceedings of the Seventh AAAI National Conference on Artificial Intelligence},
    pages = {100–104},
    numpages = {5},
    location = {Saint Paul, Minnesota},
    series = {AAAI'88}
}

@article{1decaf,
    author = {Graham, John R. and Decker, Keith S. and Mersic, Michael},
    title = {{DECAF} - A Flexible Multi Agent System Architecture},
    year = {2003},
    issue_date = {July-September 2003},
    publisher = {Kluwer Academic Publishers},
    address = {USA},
    volume = {7},
    number = {1–2},
    issn = {1387-2532},
    doi = {10.1023/A:1024120703127},
    journal = {Autonomous Agents and Multi-Agent Systems},
    month = {jul},
    pages = {7–27}
}

@inproceedings{maxsat_reinst,
    author = {Waters, Max and Padgham, Lin and Sardina, Sebastian},
    title = {Optimising Partial-Order Plans Via Action Reinstantiation},
    year = {2020},
    isbn = {9780999241165},
    booktitle = {Proceedings of the Twenty-Ninth International Joint Conference on Artificial Intelligence},
    articleno = {573},
    numpages = {9},
    location = {Yokohama, Yokohama, Japan},
    series = {IJCAI'20}
}

@article{maxsat,
    author = {Muise, Cara and Beck, J. and McIlraith, Sheila},
    year = {2016},
    month = {09},
    pages = {113-149},
    title = {Optimal Partial-Order Plan Relaxation via {MaxSAT}},
    volume = {57},
    journal = {Journal of Artificial Intelligence Research},
    doi = {10.1613/jair.5128}
}

@article{vhpop,
    author = {Simmons, Reid and Younes, H.},
    year = {2011},
    month = {06},
    pages = {}, 
    title = {{VHPOP}: Versatile Heuristic Partial Order Planner},
    volume = {20},
    journal = {Journal of Artificial Intelligence Research},
    doi = {10.1613/jair.1136}
}

@inproceedings{petri_net,
    author = {Hickmott, Sarah and Rintanen, Jussi and Thi\'{e}baux, Sylvie and White, Lang},
    title = {Planning via Petri Net Unfolding},
    year = {2007},
    publisher = {Morgan Kaufmann Publishers Inc.},
    address = {San Francisco, CA, USA},
    booktitle = {Proceedings of the 20th International Joint Conference on Artificial Intelligence},
    pages = {1904–1911},
    numpages = {8},
    location = {Hyderabad, India},
    series = {IJCAI'07}
}

@inproceedings{blackbox,
author = {Kautz, Henry and Selman, Bart},
year = {1998},
month = {June},
title = {{BLACKBOX}: A New Approach to the Application of Theorem Proving to Problem Solving},
booktitle = {AIPS98 Workshop on Planning as Combinatorial Search}
}

@inproceedings{STAN,
author = {Fox, Maria and Long, Derek},
title = {Hybrid {STAN}: Identifying and Managing Combinatorial Optimisation Sub-Problems in Planning},
year = {2001},
isbn = {1558608125},
publisher = {Morgan Kaufmann Publishers Inc.},
address = {San Francisco, CA, USA},
booktitle = {Proceedings of the 17th International Joint Conference on Artificial Intelligence - Volume 1},
pages = {445–450},
numpages = {6},
location = {Seattle, WA, USA},
series = {IJCAI'01}
}

@inproceedings{IPP,
  title={Handling of Conditional Effects and Negative Goals in {IPP}},
  author={Jana Koehler},
  year={1999}
}

@inproceedings{Veloso2002,
    author = {Veloso, Manuela and Perez, M. and Carbonell, Jaime},
    year = {2002},
    month = {02},
    title = {Nonlinear Planning with Parallel Resource Allocation},
    booktotle = {Proceedings of the Workshop on Innovative Approaches to Planning, Scheduling and Control}
}

@article{backstrom1998,
author = {Bäckström, Christer},
year = {1998},
month = {08},
pages = {99-137},
title = {Computational aspects of reordering plans},
volume = {9},
journal = {Journal of Artificial Intelligence Research}
}

@inproceedings{Kambhampati1,
author = {Nguyen, XuanLong and Kambhampati, Subbarao},
year = {2001},
month = {01},
pages = {459-466},
title = {Reviving Partial Order Planning},
journal = {International Joint Conference on Artificial Intelligence}
}

@article{KAMBHAMPATI1994235,
title = "Multi-contributor causal structures for planning: a formalization and evaluation",
author = "Subbarao Kambhampati",
year = "1994",
month = sep,
doi = "10.1016/0004-3702(94)90083-3",
language = "English (US)",
volume = "69",
pages = "235--278",
journal = "Artificial Intelligence",
issn = "0004-3702",
publisher = "Elsevier",
number = "1-2",
}

@article{kk,
title = "A unified framework for explanation-based generalization of partially ordered and partially instantiated plans",
author = "Subbarao Kambhampati and Smadar Kedar",
year = "1994",
month = may,
doi = "10.1016/0004-3702(94)90011-6",
volume = "67",
pages = "29--70",
journal = "Artificial Intelligence",
issn = "0004-3702",
publisher = "Elsevier",
number = "1",
}

@article{fd,
author = {Helmert, Malte},
year = {2011},
month = {09},
pages = {},
title = {The Fast Downward Planning System},
volume = {26},
journal = {Journal of Artificial Intelligence Research},
doi = {10.1613/jair.1705}
}

@book{lipovetzky2019introduction,
  title={An Introduction to the Planning Domain Definition Language},
  author={Haslum, Patrik and Lipovetzky, Nir and Magazzeni, Daniele and Muise, Christian},
  isbn={9781681735122},
  series={Synthesis Lectures on Artificial Intelligence and Machine Learning},
  year={2019},
  publisher={Morgan \& Claypool}
}

@inproceedings{ucpop,
    author = {Penberthy, J. Scott and Weld, Daniel S.},
    title = {{UCPOP}: A Sound, Complete, Partial Order Planner for ADL},
    year = {1992},
    isbn = {1558602623},
    publisher = {Morgan Kaufmann Publishers Inc.},
    booktitle = {Proceedings of the Third International Conference on Principles of Knowledge Representation and Reasoning},
    pages = {103–114},
    numpages = {12},
    location = {Cambridge, MA},
    series = {KR'92}
}

@article{BLUM1997281,
    title = {Fast planning through planning graph analysis},
    journal = {Artificial Intelligence},
    volume = {90},
    number = {1},
    pages = {281-300},
    year = {1997},
    issn = {0004-3702},
    doi = {10.1016/S0004-3702(96)00047-1},
    author = {Avrim L. Blum and Merrick L. Furst},
}

@article{FIKES1971189,
    title = {{STRIPS}: A new approach to the application of theorem proving to problem solving},
    journal = {Artificial Intelligence},
    volume = {2},
    number = {3},
    pages = {189-208},
    year = {1971},
    issn = {0004-3702},
    doi = {10.1016/0004-3702(71)90010-5},
    author = {Richard E. Fikes and Nils J. Nilsson},
}

@article{KAMBHAMPATI1992193,
    title = {A validation-structure-based theory of plan modification and reuse},
    journal = {Artificial Intelligence},
    volume = {55},
    number = {2},
    pages = {193-258},
    year = {1992},
    issn = {0004-3702},
    doi = {10.1016/0004-3702(92)90056-4},
    author = {Subbarao Kambhampati and James A. Hendler},
}

@INPROCEEDINGS{Regnier91completedetermination,
    author = {Pierre Regnier and Bernard Fade},
    title = {Complete determination of parallel actions and temporal optimization in linear plans of action},
    booktitle = {European Workshop on Planning, volume 522 of Lecture},
    year = {1991},
    pages = {100--111},
    publisher = {Springer-Verlag}
}

@inproceedings{winner2002analyzing,
  title={Analyzing Plans with Conditional Effects.},
  author={Winner, Elly and Veloso, Manuela M},
  booktitle={AIPS},
  pages={23--33},
  year={2002}
}

@article{Coles_Coles_Fox_Long_2021, title={Forward-Chaining Partial-Order Planning}, volume={20}, url={https://ojs.aaai.org/index.php/ICAPS/article/view/13403}, DOI={10.1609/icaps.v20i1.13403}, abstractNote={ &lt;p&gt; Over the last few years there has been a revival of interest in the idea of least-commitment planning with a number of researchers returning to the partial-order planning approaches of UCPOP and VHPOP. In this paper we explore the potential of a forward-chaining state-based search strategy to support partial-order planning in the solution of temporal-numeric problems. Our planner, POPF, is built on the foundations of grounded forward search, in combination with linear programming to handle continuous linear numeric change. To achieve a partial ordering we delay commitment to ordering decisions, timestamps and the values of numeric parameters, managing sets of constraints as actions are started and ended. In the context of a partially ordered collection of actions, constructing the linear program is complicated and we propose an efficient method for achieving this. Our late-commitment approach achieves flexibility, while benefiting from the informative search control of forward planning, and allows temporal and metric decisions to be made - as is most efficient - by the LP solver rather than by the discrete reasoning of the planner. We compare POPF with the approach of constructing a sequenced plan and then lifting a partial order from it, showing that our approach can offer improvements in terms of makespan, and time to find a solution, in several benchmark domains. &lt;/p&gt; }, number={1}, journal={Proceedings of the International Conference on Automated Planning and Scheduling}, author={Coles, Amanda and Coles, Andrew and Fox, Maria and Long, Derek}, year={2021}, month={May}, pages={42-49} }

@inproceedings{bercher_ijcai2017p68,
  author    = {Pascal Bercher and Gregor Behnke and Daniel Höller and Susanne Biundo},
  title     = {An Admissible {HTN} Planning Heuristic},
  booktitle = {Proceedings of the Twenty-Sixth International Joint Conference on
               Artificial Intelligence, {IJCAI-17}},
  pages     = {480--488},
  year      = {2017},
  doi       = {10.24963/ijcai.2017/68},
  
}

@inproceedings{Bercher_2016,
author = {Bercher, Pascal and H\"{o}ller, Daniel and Behnke, Gregor and Biundo, Susanne},
title = {More than a Name? On Implications of Preconditions and Effects of Compound {HTN} Planning Tasks},
year = {2016},
isbn = {9781614996712},
publisher = {IOS Press},
address = {NLD},
doi = {10.3233/978-1-61499-672-9-225},
booktitle = {Proceedings of the Twenty-Second European Conference on Artificial Intelligence},
pages = {225–233},
numpages = {9},
location = {The Hague, The Netherlands},
series = {ECAI'16}
}

@misc{bitmonnot2020fape,
      title={{FAPE}: a Constraint-based Planner for Generative and Hierarchical Temporal Planning}, 
      author={Arthur Bit-Monnot and Malik Ghallab and Félix Ingrand and David E. Smith},
      year={2020},
      eprint={2010.13121},
      archivePrefix={arXiv},
      primaryClass={cs.AI}
}

@inproceedings{bitmonnot:hal-01319768,
  TITLE = {{Delete-free Reachability Analysis for Temporal and Hierarchical Planning (full version)}},
  AUTHOR = {Bit-Monnot, Arthur and Smith, David E. and Do, Minh},
  URL = {https://ut3-toulouseinp.hal.science/hal-01319768},
  BOOKTITLE = {{ICAPS Workshop on Heuristics and Search for Domain-independent Planning (HSDIP)}},
  ADDRESS = {London, United Kingdom},
  HAL_LOCAL_REFERENCE = {Rapport LAAS n{\textdegree} 16167},
  YEAR = {2016},
  MONTH = Jun,
  HAL_ID = {hal-01319768},
  HAL_VERSION = {v1},
}

@article{Waters_Nebel_Padgham_Sardina_2018, 
title={Plan Relaxation via Action Debinding and Deordering}, volume={28}, 
url={https://ojs.aaai.org/index.php/ICAPS/article/view/13901}, 
DOI={10.1609/icaps.v28i1.13901}, 
number={1},
journal={Proceedings of the International Conference on Automated Planning and Scheduling}, 
author={Waters, Max and Nebel, Bernhard and Padgham, Lin and Sardina, Sebastian}, year={2018}, month={Jun.}, pages={278-287} }

@article{weld_1994,
    title={An Introduction to Least Commitment Planning},
    volume={15},
    DOI={10.1609/aimag.v15i4.1109}, 
    number={4}, 
    journal={AI Magazine}, 
    author={Weld, Daniel S.}, 
    year={1994}, 
    month={Dec.}, 
    pages={27} 
}

@inproceedings{Knoblock1994,
  title={Generating Parallel Execution Plans with a Partial-order Planner},
  author={Craig A. Knoblock},
  booktitle={International Conference on Artificial Intelligence Planning Systems},
  year={1994},
pages ={98-103}
}

@article{Boutilier_2001,
   title={Partial-Order Planning with Concurrent Interacting Actions},
   volume={14},
   ISSN={1076-9757},
   DOI={10.1613/jair.740},
   journal={Journal of Artificial Intelligence Research},
   author={Boutilier, C. and Brafman, R. I.},
   year={2001},
   month={April}, pages={105–136} }

\end{document}